\documentclass{article} 
\usepackage{iclr2026_malgai,times}
\iclrfinalcopy

\usepackage{amsmath,amsfonts,bm}

\def\eqref#1{equation~\ref{#1}}

\def\1{\bm{1}}

\DeclareMathAlphabet{\mathsfit}{\encodingdefault}{\sfdefault}{m}{sl}
\SetMathAlphabet{\mathsfit}{bold}{\encodingdefault}{\sfdefault}{bx}{n}

\usepackage{hyperref}
\usepackage{url}
\usepackage[table]{xcolor}
\usepackage[utf8]{inputenc} 
\usepackage[T1]{fontenc}    
\usepackage{url}            
\usepackage{booktabs}       
\usepackage{amsfonts}       
\usepackage{nicefrac}       
\usepackage{microtype}      
\usepackage{xcolor}         
\usepackage[textsize=tiny]{todonotes}
\usepackage{amsmath}
\usepackage{amssymb}
\usepackage{mathtools}
\usepackage{amsthm}
\usepackage{graphicx}
\usepackage{subfigure}
\usepackage{multirow}
\usepackage{tablefootnote}
\usepackage{makecell}
\usepackage{algorithmic}
\usepackage[linesnumbered,ruled]{algorithm2e}
\usepackage{wrapfig}
\usepackage{lipsum,caption}
\usepackage{pifont}
\usepackage{lipsum}
\usepackage{adjustbox}
\usepackage{tabularx}
\usepackage{enumerate}
\usepackage{etoolbox}
\usepackage{diagbox}
\usepackage{threeparttable}
\usepackage{siunitx}
\usepackage{placeins}
\usepackage{float}
\usepackage{arydshln}

\definecolor{Gray}{gray}{0.85}
\newcommand{\method}{CLEANER}

\title{
  \raisebox{-0.3em}{\includegraphics[height=1.5em]{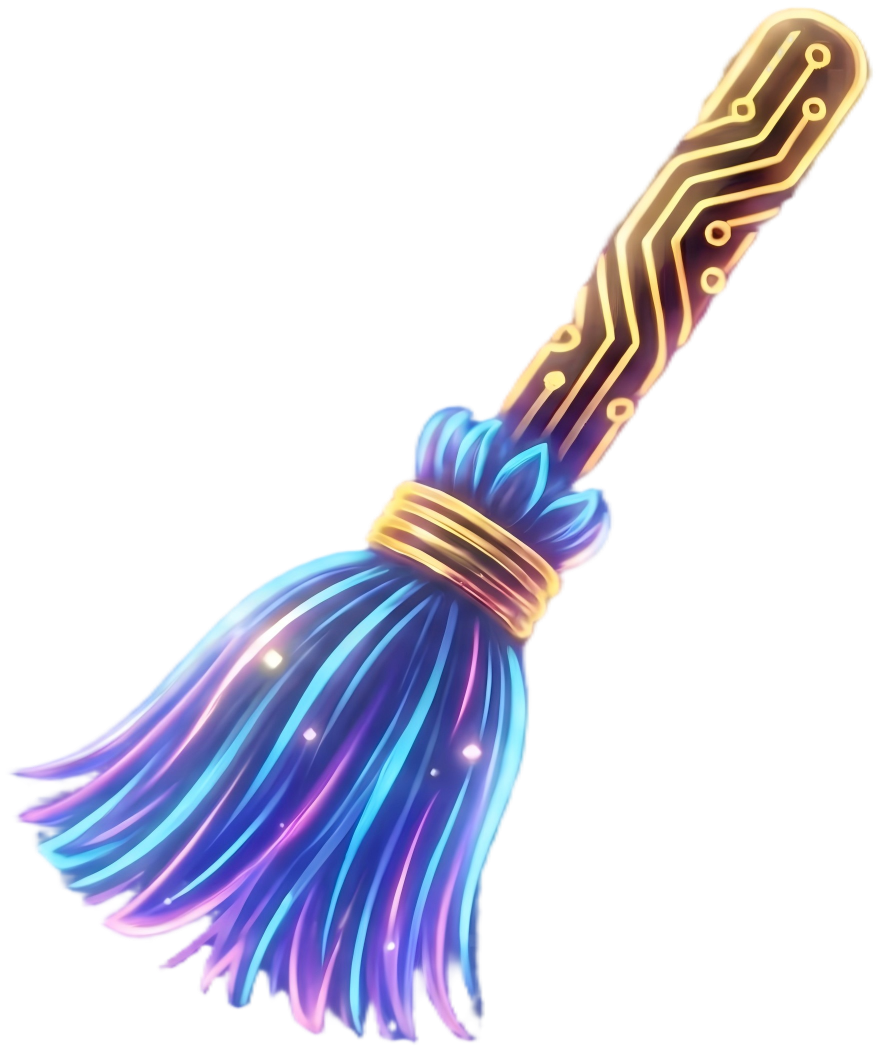}}
  \method: Self-Purified Trajectories Boost Agentic Reinforcement Learning
}

\author{
  Tianshi Xu\textsuperscript{\dag} \\
  Peking University \\
  {\small \texttt{tianshixu@stu.pku.edu.cn}} \\
  \And
  Yuteng Chen\textsuperscript{\dag} \\
  NTU, Singapore \\
  {\small \texttt{yuteng003@e.ntu.edu.sg}} \\
  \And
  Meng Li \\
  Peking University \\
  {\small \texttt{meng.li@pku.edu.cn}}
}
\begin{document}
\begingroup
\renewcommand{\thefootnote}{\fnsymbol{footnote}}
\footnotetext[2]{These authors contributed equally to this work.}
\endgroup

\maketitle

\begin{abstract}
Agentic Reinforcement Learning (RL) has empowered Large Language Models (LLMs) to utilize tools like Python interpreters for complex problem-solving. However, for parameter-constrained models (e.g., 4B--7B), the exploration phase is often plagued by frequent execution failures, creating noisy trajectories that hinder policy optimization. Under standard outcome-based reward settings, this noise leads to a critical \textit{credit assignment issue}, where erroneous actions are inadvertently reinforced alongside successful outcomes. Existing mitigations face a dilemma: dense rewards often trigger \textit{reward hacking}, while supersampling incurs {prohibitive computational costs}. To address these challenges, we propose \textbf{\method{}}. Distinct from external filtering methods, \method{} exploits the model's {intrinsic self-correction capabilities} to eliminate error-contaminated context directly during data collection. At its core, the \textbf{Similarity-Aware Adaptive Rollback (SAAR)} mechanism autonomously constructs clean, purified trajectories by retrospectively replacing failures with successful self-corrections. Based on semantic similarity, SAAR adaptively regulates replacement granularity from shallow execution repairs to deep reasoning substitutions. By training on these self-purified paths, the model internalizes correct reasoning patterns rather than error-recovery loops. Empirical results on AIME24/25, GPQA, and LiveCodeBench show average accuracy gains of \textbf{6}\%, \textbf{3}\%, and \textbf{5}\% over baselines. Notably, \method{} matches state-of-the-art performance using only \textbf{one-third} of the training steps, highlighting trajectory purification as a scalable solution for efficient agentic RL. Our models and code are available at \href{https://github.com/Tianshi-Xu/Open-CLEANER}{GitHub}.
\end{abstract}
\begin{figure}[H]
    \centering
    \includegraphics[width=0.8\linewidth]{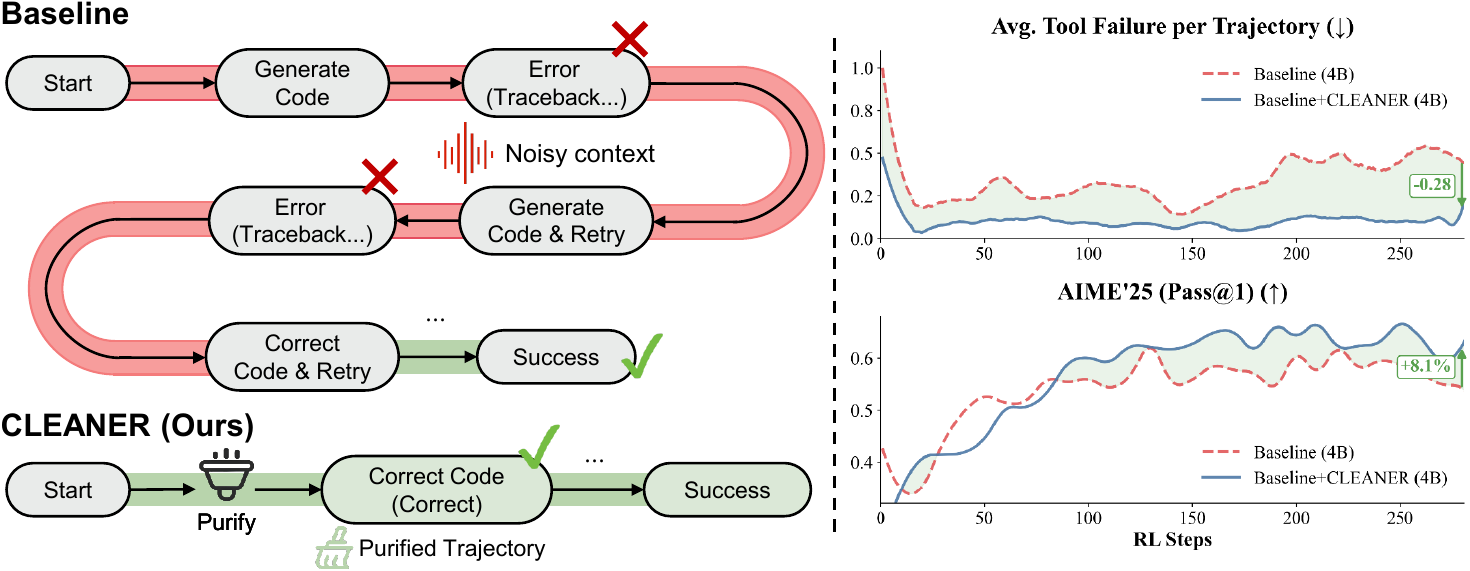}
    \caption{\textbf{Left}: Illustration of the differences between the standard baseline and our \method{}.
    \textbf{Right}: By reducing the number of tool execution failures within trajectories during training, our method improves pass@1 accuracy on AIME’25 by 8.1\%.}
    \label{fig:intro}
\end{figure}
\section{Introduction}
The landscape of Large Language Models (LLMs) is shifting from passive text generation systems toward autonomous agents that solve complex tasks through tool use~\citep{yao2022react,gou2023tora, schick2023toolformer,jin2025search,li2025torl,feng2025retool}. Among the diverse tool modalities available to LLM agents, the Python code interpreter plays a particularly critical role~\cite{wang2024executable,shang2025rstar2,yu2025demystifying}. Due to its Turing completeness and deterministic execution semantics, Python is indispensable for tasks that require precise computation, including mathematical reasoning, algorithmic problem-solving, and data analysis. As observed by Ronacher~\citep{ronacher2025code}, code is increasingly serving as a ``universal interface'' that unifies logical reasoning, computation, and API interaction within a single expressive medium. This perspective aligns with frameworks like CodeAct~\citep{wang2024executable}, which advocate for treating code execution as a first-class action in agentic reasoning. By doing so, agents can effectively plan, verify intermediate results, and iteratively correct errors through interaction with an execution environment. Together, these insights highlight that robust code synthesis and execution are foundational capabilities for tool-augmented LLM agents.
Motivated by this perspective, this work focuses on \emph{Python code execution} as the primary tool modality. 

However, fully realizing this potential presents significant challenges for parameter-constrained models (e.g., 4B–7B). A primary obstacle is the high rate of execution failure, particularly during the exploration phase of Reinforcement Learning (RL)~\citep{guo2025deepseekr1}. Before policy convergence, these models frequently generate invalid code, causing the intended recovery mechanism to degenerate into prolonged ``Error $\rightarrow$ Feedback $\rightarrow$ Retry'' loops, as depicted in Figure~\ref{fig:intro} (left).
This instability constitutes a critical bottleneck in training. As evidenced by the experimental results in Figure~\ref{fig:intro} (right), an excessive accumulation of tool errors within trajectories closely correlates with performance bottlenecks or even accuracy degradation. We attribute this to the pollution of context: repeated failures generate large amounts of misleading signals (e.g., invalid code and verbose tracebacks). This accumulated noise likely causes {semantic interference}, biasing the model toward rationalizing incorrect execution paths rather than re-grounding its decisions, thereby hindering policy improvement.

In principle, RL algorithms are expected to guide models away from such instability~\citep{yu2025dapo,chen2025minimax}. However, standard training paradigms frequently worsen the problem due to the \textbf{credit assignment issue}. Under sparse, outcome-based reward settings like GRPO~\cite{guo2025deepseekr1}, the entire trajectory receives a uniform positive reward upon final success, regardless of preceding failures. This mechanism fails to distinguish between efficient reasoning and trajectories containing errors, effectively treating them as equivalent. Consequently, erroneous tool usage and the underlying logic are inadvertently reinforced despite their negative impact on reasoning.

To mitigate this, prior research has explored various strategies, yet each introduces new flaws. Attempts to assign dense rewards for individual tool executions often suffer from \textit{reward hacking}, biasing agents toward optimizing intermediate metrics rather than final outcomes~\cite{yu2025demystifying}. Alternatively, works such as \emph{rstar2-agent}~\citep{shang2025rstar2} utilize {supersampling-based trajectory filtering}, retaining only high-quality instances from $2\times$ generated candidates. However, this incurs a prohibitive computational cost. Since the rollout phase usually dominates RL training (accounting for $>80\%$ of runtime~\citep{li2023remax,sheng2024hybridflow,fu2025areal}), such extensive sampling renders these strategies unscalable for resource-constrained settings.

To address these challenges, we propose \textbf{\method{}} (\emph{Self-Purified Trajectories Boost Agentic Reinforcement Learning}). \method{} significantly boosts the agentic RL by eliminating error-contaminated context from the training data. Unlike methods that rely on increasing rollout multiplicity, \method{} operates specifically at the data level to refine the trajectories used for policy optimization. At the core of our approach is the \textbf{Similarity-Aware Adaptive Rollback (SAAR)} mechanism, which constructs self-purified trajectories. When the model generates incorrect code but subsequently self-corrects within the same rollout, SAAR intervenes to prevent the error-laden history from being used for optimization. Instead, it applies a retrospective context substitution where the trajectory is rolled back to the failure point and the erroneous action is replaced with the corrected solution. This process yields a revised trajectory containing substantially fewer execution errors.
To ensure semantic coherence, SAAR adaptively regulates the rollback granularity based on the semantic similarity between the erroneous code and its corrected counterpart. High-similarity cases typically correspond to minor execution errors and trigger a shallow replacement that preserves the original reasoning. Conversely, low-similarity cases signal deeper logical flaws and necessitate the substitution of the entire reasoning segment to maintain consistency. By leveraging these self-purified trajectories, \method{} reduces noise in the learning signal and accelerates capability acquisition. Empirical evaluations show that \method{} outperforms standard baselines with average accuracy gains of approximately 6\% on AIME, 3\% on GPQA~\cite{rein2024gpqa}, and 5\% on LiveCodeBench~\cite{jain2024livecodebench}. Furthermore, it matches the performance of state-of-the-art (SOTA) models~\cite{yu2025demystifying} while requiring only one-third of the RL steps.
In summary, our main contributions are as follows:
\begin{enumerate}[9]
    \item[\ding{182}] We propose \textbf{\method{}}, which resolves the credit assignment dilemma in agentic RL by training on \emph{self-purified trajectories}. This approach enables models to directly internalize correct reasoning patterns while filtering out the interference of execution noise.
    
    \item[\ding{183}] We introduce the \textbf{SAAR} mechanism to autonomously construct these clean signals. SAAR adaptively repairs failures—ranging from minor syntax typos to deep logical flaws—without the computational overhead of supersampling.
    
    \item[\ding{184}] We demonstrate that \method{} achieves state-of-the-art efficiency and performance. It outperforms baselines with accuracy gains of 6\% on AIME and 5\% on LiveCodeBench, and notably matches SOTA performance using only \textbf{one-third} of the training steps.

    \item[\ding{185}] We provide a fully reproducible training pipeline and have made our code, environment configurations, and processed datasets available via \href{https://anonymous.4open.science/r/Open-CLEANER-anonymous--B6E6}{Anonymous GitHub} to support further research.
\end{enumerate}
\section{Preliminaries}
\subsection{Agentic Reasoning Trajectories}
\noindent \textbf{Notation.} We formalize the agent’s problem-solving process as a sequential generation task over a growing trajectory history. Let $\mathcal{M}$ denote the large language model acting as the agent, and let $\mathcal{E}$ denote the code execution environment (i.e., a Python interpreter).
At turn $t$, the interaction history is denoted by $h_t$, which consists of the initial user query $x$ and a sequence of past interaction tuples:
\begin{equation}
h_t = \bigl[ x,\,(r_0, c_0, o_0), \dots, (r_{t-1}, c_{t-1}, o_{t-1}) \bigr].
\end{equation}
For each turn $i$, $r_i$ denotes the \emph{reasoning trace} expressed in natural language, $c_i$ denotes the \emph{code action} corresponding to an executable Python program, and $o_i$ denotes the \emph{observation} returned by the execution environment, i.e., $o_i = \mathcal{E}(c_i)$. We distinguish between successful executions, denoted by $o_i^{+}$, and execution failures or runtime errors, denoted by $o_i^{-}$.

\noindent \textbf{Standard Generation Process.}
At step $t$, the policy $\pi_{\theta}$ conditions on the current history $h_t$ and generates a reasoning trace and a code action:
\begin{equation}
(r_t, c_t) \sim \pi_{\theta}(\cdot \mid h_t).
\end{equation}
The environment then executes the generated code and returns an observation $o_t = \mathcal{E}(c_t)$. The interaction history is updated by appending the new tuple:
\begin{equation}
h_{t+1} = h_t \oplus (r_t, c_t, o_t),
\end{equation}
where $\oplus$ denotes sequence concatenation. In standard training pipelines, execution failures ($o_t^{-}$) are permanently recorded in the history, thereby introducing error-induced noise into subsequent conditioning and the resulting learning signal.

\subsection{Group-Based Policy Optimization with Outcome-only Reward}
We employ \emph{sparse, outcome-only supervision}, rewarding agents solely upon trajectory completion. This avoids manual reward engineering and prevents ``reward hacking''~\citep{shang2025rstar2}. For policy optimization, we use Group Relative Policy Optimization~\citep{guo2025deepseekr1} to manage credit assignment via group-relative comparisons. See Appendix~\ref{app:grpo} for mathematical formulations.

\section{Problem Formulation: Impact of Code Tool Execution Noise}
Unlike internal Chain-of-Thought reasoning, agentic workflows introduce external stochasticity via interactions with the environment $\mathcal{E}$. This uncertainty manifests as trajectory-level noise that hinders efficient policy optimization.

\noindent \textbf{Context Contamination from Erroneous Tool Calls.}
Code execution is inherently error-prone, and failed executions
($o_t = o^-$) frequently occur during exploration. While these error traces contribute minimally to the final task resolution, they are permanently appended to the trajectory history $h_t$. Consequently, these low-information segments consume valuable context window capacity and disrupt the logical flow of subsequent reasoning, effectively contaminating the agent's decision context with noise.
\begin{wrapfigure}[16]{r}[0em]{0.46\textwidth}
    \vspace{-5pt}
    \centering
    \includegraphics[width=1.0\linewidth]{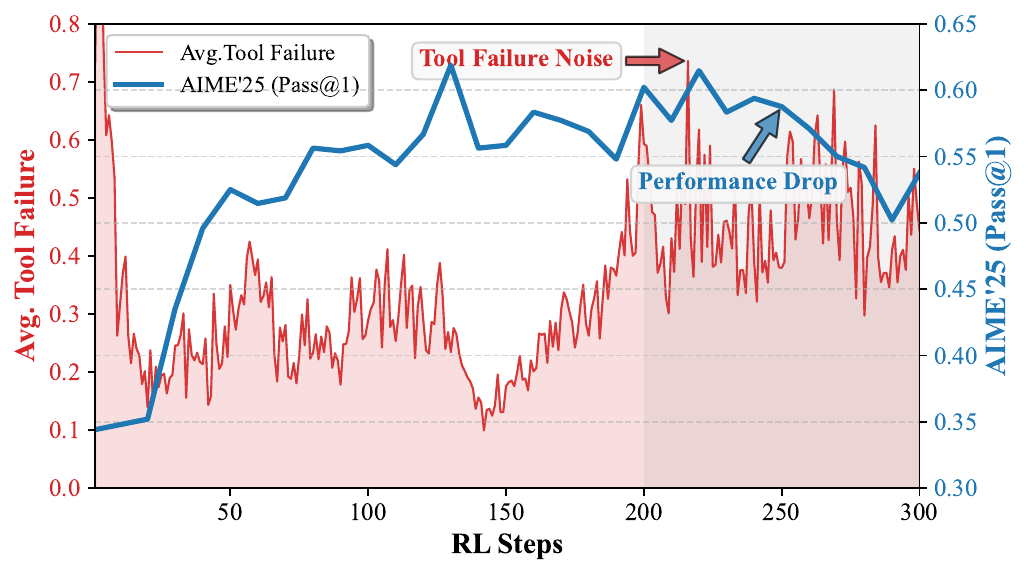}
    \caption{Impact of execution noise. Tool failure spikes correlate with AIME25 performance drops, highlighting optimization sensitivity to noisy trajectories.}
    \label{fig:problem}
\end{wrapfigure}
\noindent \textbf{Credit Assignment Ambiguity under Outcome-Only Reward.}
This uncertainty becomes particularly detrimental under sparse, outcome-based RL, where rewards are assigned solely based on final task success. As illustrated in Figure~\ref{fig:problem}, we observe a distinct phenomenon: \emph{bursts of erroneous tool calls within individual trajectories}. During these periods, accuracy plateaus or even degrades, indicating an optimization bottleneck.
The root cause of this inefficiency lies in a fundamental \textbf{credit assignment failure} within \emph{noisy successes}—trajectories that eventually succeed despite containing intermediate errors. Consider a typical noisy trajectory $\tau_{\text{noisy}}$, in which the agent initially produces incorrect code but later self-corrects:
\begin{equation}
\tau_{\text{noisy}} =
[\dots, h_t,
\underbrace{(r_t, c_{\text{err}}, o^-)}_{\text{Noise (Trial 1)}},
\underbrace{(r'_{\text{aux}}, c_{\text{corr}}, o^+)}_{\text{Signal (Trial 2)}},
\dots]
\end{equation}
Here, the agent first emits an erroneous code action $c_{\text{err}}$, receives a runtime error $o^-$, and subsequently generates a corrected code $c_{\text{corr}}$ accompanied by an auxiliary reasoning trace $r'_{\text{aux}}$, which executes successfully.
Since the reward function $R(\tau)$ is binary and episodic, the final positive reward is uniformly propagated across the entire trajectory $\tau_{\text{noisy}}$. Consequently, both the erroneous action $c_{\text{err}}$ and the corrective action $c_{\text{corr}}$ receive identical positive reinforcement, despite their fundamentally conflicting semantic roles. We refer to this effect as \textbf{Trajectory Noise}: spurious credit assigned to intermediate failures that dilutes the learning signal. Over time, this noise implicitly validates suboptimal tool usage patterns, amplifies variance in policy updates, and leads to brittle optimization.


\section{Method: \underline{S}imilarity-\underline{A}ware \underline{A}daptive \underline{R}ollback (SAAR)}
\begin{figure*}[!tb]
    \centering
    \includegraphics[width=0.9\linewidth]{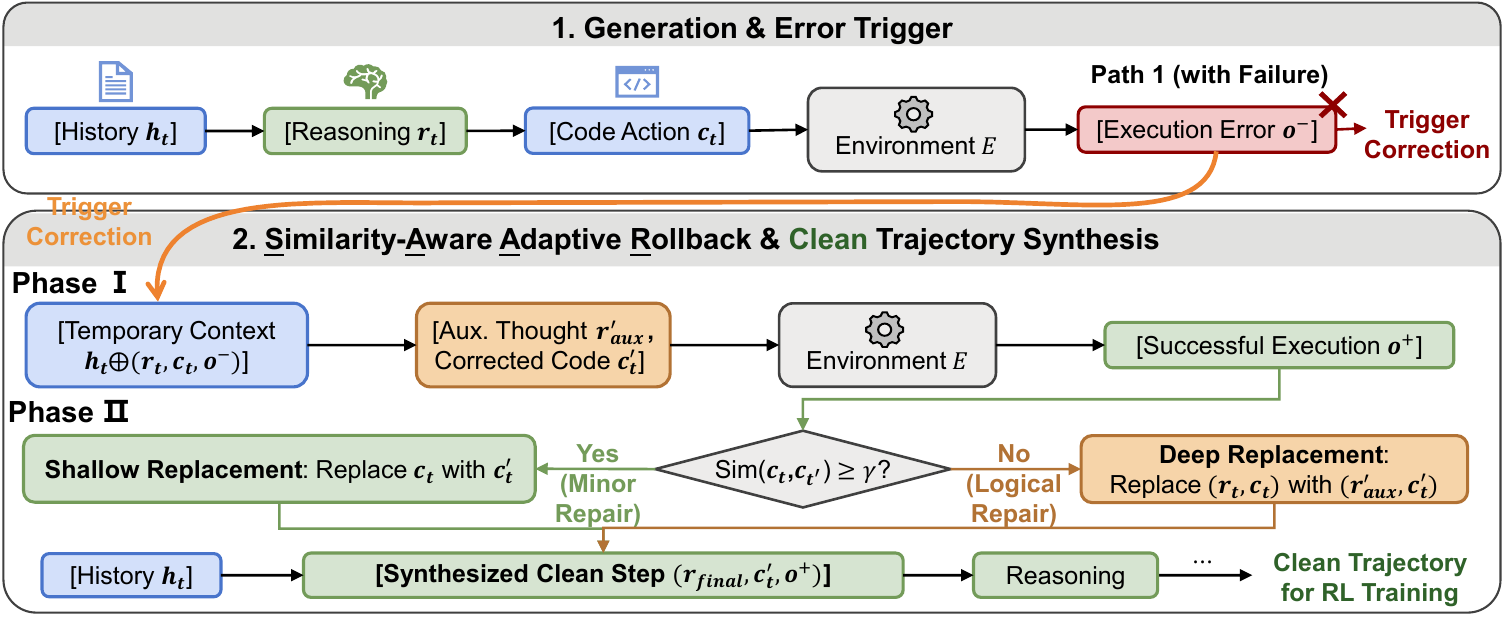}
    \caption{Illustration of our \underline{S}imilarity-\underline{A}ware \underline{A}daptive \underline{R}ollback (SAAR).}
    \label{fig:method}
\end{figure*}
To boost agentic reinforcement learning under noisy tool interactions, we propose \method{}, a trajectory purification framework centered on \emph{Similarity-Aware Adaptive Rollback (SAAR)}. The core objective is to distill the learning signal by retrospectively eliminating execution failures from exploration rollouts, thereby constructing \emph{clean, self-purified trajectories}. In these synthesized paths, the agent appears to solve the task fluently, enabling the optimizer to reinforce correct reasoning logic rather than error-recovery loops. As illustrated in Figure~\ref{fig:method}, this data-level intervention is triggered by execution errors and operates through a two-phase process:
\subsection{Phase I: Error Trigger and Lookahead Correction}

At time step $t$, when the environment returns an execution error $o_t^-$ following code action $c_t$, we defer committing this failure to the history. Instead, we freeze the current state $h_t$ and initiate a temporary lookahead phase to seek a viable solution.

\noindent \textbf{Context Extension.}
We temporarily construct an augmented context that exposes the execution error, allowing the model to analyze the feedback:
\begin{equation}
\tilde{h}_t = h_t \oplus (r_t, c_t, o_t^-).
\end{equation}

\noindent \textbf{Correction Generation.}
Conditioned on $\tilde{h}_t$, the policy generates a corrective response, typically comprising an auxiliary reasoning trace $r'_{\text{aux}}$ and a revised code action $c'_t$:
\begin{equation}
r'_{\text{aux}}, c'_t \sim \pi_\theta(\cdot \mid \tilde{h}_t).
\end{equation}

\noindent \textbf{Verification.}
The revised code is executed to obtain a new observation $o'_t = \mathcal{E}(c'_t)$. If execution succeeds ($o'_t = o^+$), we proceed to Phase II to integrate this success into the trajectory. If failure persists, the correction loop repeats up to $K$ attempts.

\subsection{Phase II: Similarity-Aware Adaptive Replacement}

Upon obtaining a valid correction $c'_t$, SAAR determines the optimal strategy to merge it into the history $h_t$. The intuition is that the semantic distance between the error $c_t$ and correction $c'_t$ reveals the nature of the failure. We quantify this using a similarity function $\mathrm{Sim}(c_t, c'_t)$, implemented via $\operatorname{difflib.SequenceMatcher}$, and compare it against a code similarity threshold $\gamma$.

\noindent \textbf{Case A: Implementation-Level Repair ($\mathrm{Sim}(c_t, c'_t) \ge \gamma$).}
High similarity indicates a superficial error (e.g., syntax typos), where the original reasoning $r_t$ is presumed to be sound. In this scenario, we perform a shallow replacement: the failed action $c_t$ and error $o_t^-$ are discarded, and the corrected code $c'_t$ is grafted directly onto the existing reasoning $r_t$. This yields the purified tuple $(r_t, \mathbf{c'_t}, o'_t)$.

\noindent \textbf{Case B: Reasoning-Level Repair ($\mathrm{Sim}(c_t, c'_t) < \gamma$).}
Low similarity signals a substantial divergence in implementation strategy, suggesting that the initial reasoning $r_t$ is likely incompatible or misaligned with the corrected solution. Retaining the outdated reasoning would introduce semantic dissonance within the training data. Thus, we execute a \emph{deep replacement}: the entire failed turn $(r_t, c_t, o_t^-)$ is removed, and the auxiliary correction thought $r'_{\text{aux}}$ is adopted as the canonical reasoning, forming the consistent tuple $(\mathbf{r'_{\text{aux}}}, \mathbf{c'_t}, o'_t)$.

Through this adaptive mechanism, we synthesize the \emph{self-purified trajectory}:
\begin{equation}
\tau_{\text{purified}} =
[\dots, h_t,
\underbrace{( \mathbf{r_{\text{final}}}, \mathbf{c'_t}, o'_t )}_{\text{Purified Context}},
\dots],
\end{equation}
where $r_{\text{final}} \in \{ r_t, r'_{\text{aux}} \}$ is determined by the rollback granularity. This constructs a coherent, counterfactual history of immediate success, effectively guiding the policy to internalize correct reasoning patterns while bypassing the noise of trial-and-error.

\subsection{Implementation Details}

\noindent \textbf{Logit Recomputation via RadixAttention.}
Since the corrected action $c'_t$ is sampled from the error-augmented context $\tilde{h}_t$, there exists a distribution shift: $\pi_\theta(c'_t \mid \tilde{h}_t) \neq \pi_\theta(c'_t \mid h_t\oplus r_{final})$. To ensure the policy update is grounded in the correct causal path, we must recompute the log-probabilities of $c'_t$ under the purified context. To minimize overhead, we employ SGLang~\citep{zheng2024sglang} with \emph{RadixAttention}. This mechanism efficiently reuses the KV cache for the invariant history prefix, restricting the computational cost strictly to the modified suffix segments.

\noindent \textbf{Curriculum Mixing for Robustness.} 
To balance error avoidance with error recovery, we employ a stochastic mixing strategy for Qwen2.5-7B. Specifically, we apply SAAR to 70\% of trajectories while retaining 30\% in their original state. This curriculum ensures the model internalizes correct reasoning patterns without sacrificing its intrinsic ability to debug and self-correct during inference.
\section{Experiments}
\subsection{Experimental Setup}
\begin{figure*}[!tb]
    \centering
    \includegraphics[width=1.0\linewidth]{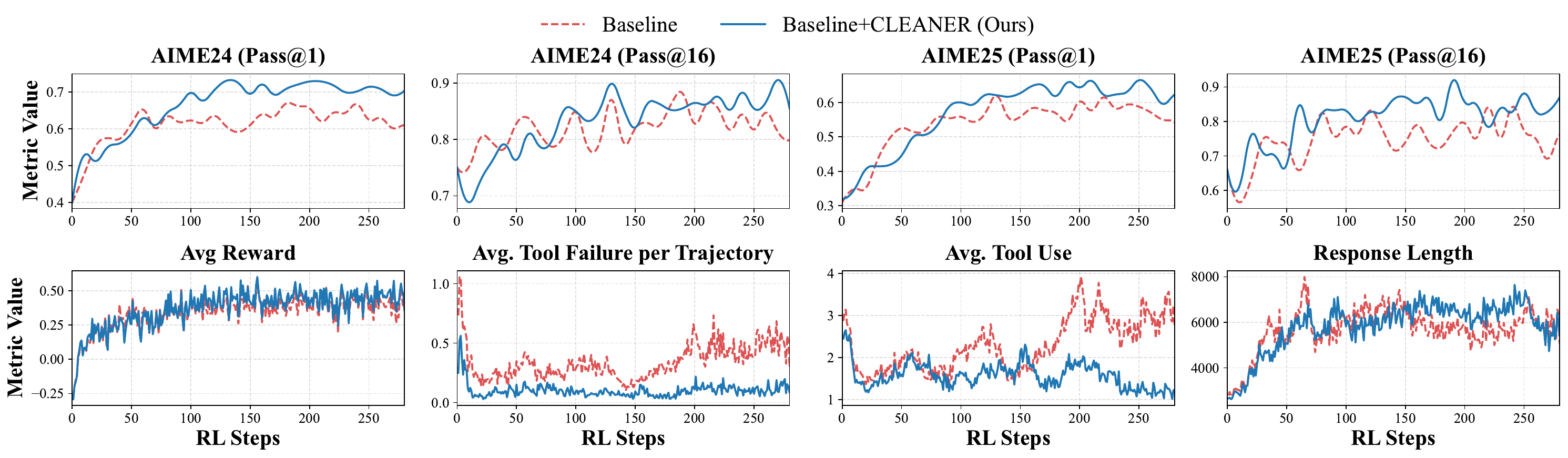}
    \caption{\textbf{Evolution of training metrics during RL.} Compared to the DAPO-baseline, \method{} effectively suppresses erroneous tool calls in trajectories, leading to significant performance gains.}
    \label{fig:exp}
\end{figure*}
\begin{table*}[!tb]
\centering
\caption{\textbf{Comparing \method{} with existing works.} \textbf{Bolded} entries denote the top-performing methods initialized from the same Qwen3-4B-Instruct base. Despite its compact 4B scale, \method{} matches the performance of significantly larger models and achieves results {comparable to SOTA baselines} while requiring only \textbf{one-third} of the training steps utilized by DemyAgent-4B.}
\label{tab:main_results}\label{tab:main_results}
\resizebox{0.88\textwidth}{!}{
\begin{tabular}{lcccccc}
\toprule
\multicolumn{1}{c}{\multirow{3}{*}{\textbf{Method}}} & \multicolumn{2}{c}{\textbf{MATH}} & \textbf{Science} & \multicolumn{2}{c}{\textbf{LiveCodeBench}} & \multirow{3}{*}{\makecell{\textbf{RL Step}\\(Batch Size=128)} }\\
\cmidrule(lr){2-3} \cmidrule(lr){4-4} \cmidrule(lr){5-6}
 & \makecell{AIME24} & \makecell{AIME25} & GPQA & V6 & Whole \\
\midrule
\rowcolor{gray!20}
\multicolumn{7}{l}{\textit{Self-Contained Reasoning}} \\
 Qwen2.5-7B-Instruct & 16.7 & 10.0 & 31.3 &15.2 & - &/  \\
 Qwen3-4B-Instruct-2507 & 63.3 & 47.4 & 52.0 &35.1 & - &/  \\
 Qwen2.5-72B-Instruct & 18.9 & 15.0 & 49.0& - & - & /\\
 DeepSeek-V3 &39.2&28.8&59.1& 16.1 & 49.6 &/ \\
 DeepSeek-R1-Distill-32B & 70.0 & 46.7 & 46.7 & -&- & / \\
 DeepSeek-R1-Zero (671B) & 71.0 & 53.5 & 53.5&- &-& / \\
\midrule
\rowcolor{gray!20}
\multicolumn{7}{l}{\textit{Agentic Reasoning}} \\
 ToRL-7B & 43.3 & 30.0 & - &- &-&550 \\
 ReTool-32B & 72.5 & 54.3 & - &-&-& 1200 \\
 Tool-Star-3B & 20.0 & 16.7 & - &-&-& 120 \\
 ARPO-7B & 30.0 & 30.0 & 53.0 &12.1&15.8 &157 \\
 AEPO-7B & 33.0 & 30.0 & 55.6 & 14.3&17.8 & 157\\
 rStar2-Agent-14B & 80.6 & 69.8 & 60.9&- &-& 500 \\
 \cdashline{1-7}
 \addlinespace
 DemyAgent-4B (Qwen3-4B-Instruct) & 72.6 & \textbf{70.0} & 58.5 &26.8&51.7 & 750 \\
 DAPO-baseline (Qwen3-4B-Instruct) & 66.7 & 59.4 & 56.9 & 26.6 & 49.5 & 250 \\
 \textbf{\method-4B (Qwen3-4B-Instruct)} &  \textbf{72.7} & 67.1 &  \textbf{60.2} &  \textbf{26.8}& \textbf{54.9}& \textbf{250} \\
\bottomrule
\end{tabular}
}
\end{table*}
\noindent \textbf{Models and Training Datasets.} 
We evaluate our framework using Qwen3-4B-Instruct-2507~\citep{yang2025qwen3} and Qwen2.5-7B-Instruct~\citep{yang2024qwen2}. The models first undergo cold-start Supervised Fine-Tuning (SFT) followed by reinforcement learning (RL), both utilizing the agentic datasets provided by \citep{yu2025demystifying}. Comprehensive dataset specifications and processing details are documented in Appendix~\ref{app:hyper}.

\noindent \textbf{Implementation.}
We implement our training pipeline using the {VeRL} framework~\citep{sheng2024hybridflow} distributed via {PyTorch FSDP2}. We employ the code judge from~\citep{shang2025rstar2} as the Python interpreter, which ensures robust stability even under the heavy concurrency of tool invocations during the RL rollout phase. Additionally, our prompt design adheres to the specifications outlined in~\citep{yu2025demystifying}. Trajectory rollouts are generated using {SGLang}~\citep{zheng2024sglang}. 
To address severe training instability caused by numerical inconsistencies between the training and rollout phases, we adopt FP16 precision for rollout generation following \citep{qi2025defeatingfp16}.

\noindent \textbf{Training Recipe.}
We train the models for one epoch using the DAPO algorithm~\cite{yu2025dapo}. We employ a rollout batch size of 128, a group size of 16, and an update mini-batch size of 32. The learning rate is set to $2\text{e-}6$ for the 4B model and $1\text{e-}6$ for the 7B model. Specifically for the Qwen2 experiments, we generate 8 rollouts per query and filter out instances that are either trivially easy or unsolvable to ensure stability. Further experimental details are provided in Appendix~\ref{app:hyper}.

\noindent \textbf{Evaluation Benchmarks.} To comprehensively demonstrate the improvements in reasoning and coding capabilities achieved by our method, we conduct evaluations across four challenging benchmarks: AIME24, AIME25, LiveCodeBench~\citep{jain2024livecodebench}, and GPQA~\citep{rein2024gpqa}. To ensure a fair comparison, all models are evaluated using identical sampling parameters; detailed specifications are provided in the Appendix~\ref{app:hyper}.

\noindent \textbf{Baselines.}
To rigorously assess the efficacy of \method{}, we compare it against baselines across two distinct categories: \textbf{\underline{1)}} \textit{Self-Contained Reasoning.} We include standard instruction-tuned and reasoning-specialized models: Qwen2.5-7B-Instruct~\citep{hui2024qwen2}, Qwen3-4B-Instruct-2507~\citep{yang2025qwen3}, Qwen2.5-72B-Instruct, DeepSeek-V3~\citep{liu2024deepseek}, DeepSeek-R1-Distill-32B, and DeepSeek-R1-Zero (671B)~\citep{guo2025deepseekr1}.
\textbf{\underline{2)}}\textit{ Agentic Reasoning.} We compare against SOTA agentic models, including ToRL-7B~\citep{li2025torl}, ReTool-32B~\citep{feng2025retool}, Tool-Star-3B~\citep{dong2025toolstar}, ARPO-7B~\citep{dong2025arpo}, AEPO-7B~\citep{dong2025aepo}, Demystify-4B~\citep{yu2025demystifying}, and rStar2-Agent-14B~\citep{shang2025rstar2}. We also include a DAPO-baseline that shares an identical training configuration with \method{}, with the sole exception of excluding the SAAR mechanism. 


\subsection{Main Results}
\textbf{\method{} Converts Execution Noise into Effective Reasoning.}
Figure~\ref{fig:exp} illustrates the evolution of key metrics for both the DAPO-baseline and \method{} during RL. Empirical results support three primary conclusions: 
\textbf{\underline{1)}} \textbf{Error Suppression:} Through SAAR, \method{} consistently suppresses erroneous tool calls to a minimal level, mitigating interference with the model's reasoning process. 
\textbf{\underline{2)}} \textbf{Performance Gains:} Reduced noise translates to significant improvements on AIME24/25 (avg. +6\% Pass@1, +8\% Pass@16), demonstrating enhanced exploration.
\textbf{\underline{3)}} \textbf{Efficient Reasoning:} Despite comparable output lengths, the reduction in errors implies that \method{} reallocates tokens from futile tool calls to effective reasoning, facilitating deeper thinking.

\noindent \textbf{Compared to Previous Works.}
The main results comparing \method{} with existing works are summarized in Table~\ref{tab:main_results}. We observe the following: 
\textbf{\underline{1)}} Compared to Self-Contained Reasoning models that lack specialized training for agentic scenarios, \method{} demonstrates robust performance despite its compact 4B parameter size. This validates that small models, when subject to tailored post-training, can achieve capabilities comparable to significantly larger counterparts. 
\textbf{\underline{2)}} In contrast to the SOTA baseline DemyAgent-4B, \method{} attains comparable results using only \textbf{one-third} of the training steps. Notably, it surpasses it on AIME24, GPQA, and LiveCodeBench. We attribute this efficiency to the purified trajectories, which enable the model to acquire coding and reasoning capabilities more rapidly and effectively. Conversely, the DAPO-baseline exhibits significantly lower accuracy under limited training (250 steps), due to interference from tool call noise.

\subsection{Ablation Study}
\begin{wraptable}[14]{r}{0.50\textwidth}
    \centering
    \normalsize
    \caption{Ablation study on the effectiveness of \method{}.}\label{tab:exp_ablation}
    \resizebox{\linewidth}{!}{
        \setlength{\tabcolsep}{2pt} 
        \begin{tabular}{lcccc}
        \toprule
        \multicolumn{1}{c}{\textbf{Method}} & \makecell{AIME24 \\ Pass@1/16} & \makecell{AIME25 \\ Pass@1/16} & GPQA & LiveCodeBench-v6 \\
        \midrule
        \rowcolor{gray!20}
        \multicolumn{5}{l}{\textit{Qwen3-4B-Instruct}} \\
         RL w/o Tools & 64.0/78.8 & 53.3/77.0 & 52.2 & 19.8\\
         + Tools & 66.7/84.4 & 59.4/84.2 & 56.9& 26.6 \\
         + SAAR (Ours) & \textbf{72.7/87.6}& \textbf{67.1}/84.1& \textbf{60.2} &  \textbf{26.8} \\
         \rowcolor{gray!20}
        \multicolumn{5}{l}{\textit{Qwen2.5-7B-Instruct}} \\
         RL w/o Tool & 15.4/30.5 & 14.4/24.4  & 32.3 & 1.1\\
         + Tools & 40.2/59.1 & 27.3/46.3  & 35.9 & 13.0\\
         + SAAR (Ours)& \textbf{44.6/64.3} & \textbf{31.0/54.7}&  \textbf{40.0} &  \textbf{13.1} \\
        \bottomrule
        \end{tabular}
    }
\end{wraptable}
\textbf{Ablation on the Effectiveness of CLEANER.} 
As detailed in Table~\ref{tab:exp_ablation}, we evaluate three configurations under identical hyperparameters to isolate the contribution of each component: (1) \textbf{RL w/o Tools}, relying solely on internal reasoning; (2) \textbf{RL w/ Tools}, which integrates a Python code interpreter; and (3) \textbf{RL w/ Tools + SAAR} (i.e., \method{}). 
The results yield two key observations: \textbf{\underline{1)}} \textbf{The necessity of tool integration.} Equipping the model with a code interpreter significantly enhances performance on mathematical and coding tasks, improving average accuracy by over 5\% on Qwen3-4B and 20\% on Qwen2.5-7B. \textbf{\underline{2)}} \textbf{The superiority of purified trajectories.} \method{} consistently outperforms the baselines across all benchmarks and model scales. Specifically, for Qwen3-4B, we achieve average gains of 6\% on AIME, 4\% on GPQA, and 5\% on LiveCodeBench. Similarly, Qwen2.5-7B exhibits an average improvement of 4\%. These findings confirm that the trajectory purification effectively amplifies the potential of Agentic RL.
\begin{table*}[!tb]
    \centering
    \begin{minipage}{0.35\textwidth}
        \centering
        \caption{Ablation on learning rate.}\label{tab:exp_lr}
        \resizebox{\linewidth}{!}{
            \begin{tabular}{lcc}
            \toprule
            \multicolumn{1}{c}{\textbf{Method}}
             & \makecell{AIME24 \\ Pass@1/16} & \makecell{AIME25 \\ Pass@1/16} \\ 
            \midrule
            RL w/ Tools (1e-6) & 66.9/85.1 & 63.3/83.9 \\
            \method-4B (1e-6) & 70.8/85.4 & 64.2/86.4 \\
            \method-4B (2e-6) & {72.7/87.6}& {67.1}/84.1\\
            \bottomrule
            \end{tabular}
        }
    \end{minipage}%
    \hfill 
    \begin{minipage}{0.62\textwidth}
        \centering
        \caption{Ablation on SAAR deactivation. Performance and the evaluation time comparison with vs. without SAAR during the evaluation phase.}\label{tab:exp_ablation_w_o_saar}
        \resizebox{\linewidth}{!}{
            \begin{tabular}{lccccc}
            \toprule
            \multicolumn{1}{c}{\textbf{Method}} & \makecell{AIME24 \\ Pass@1/16} & \makecell{AIME25 \\ Pass@1/16} & GPQA & LiveCodeBench-v6 & \makecell{Time \\ (min)} \\
            \midrule
            \method-4B & 72.7/87.6& 67.1/84.1& 60.2 &26.8 & 115\\
            \method-4B w/o SAAR & 72.1/86.3& 64.6/84.3& 59.8 &26.6 & 106\\
            \bottomrule
            \end{tabular}
        }
    \end{minipage}
\end{table*}

\noindent \textbf{Ablation on Learning Rate.}
Table~\ref{tab:exp_lr} summarizes our ablation study on learning rates across different model scales. For the 4B model, we adopted a relatively large learning rate of $2\text{e-}6$ to accelerate convergence. As summarized in Table~\ref{tab:exp_lr}, \method{} achieves consistent improvements across different settings, with $2\text{e-}6$ yielding superior results. For the 7B model, we adopted a learning rate of $1\text{e-}6$ to ensure optimization stability for the larger parameter space.
\begin{figure*}[!tb]
    \centering
    \includegraphics[width=1.0\linewidth]{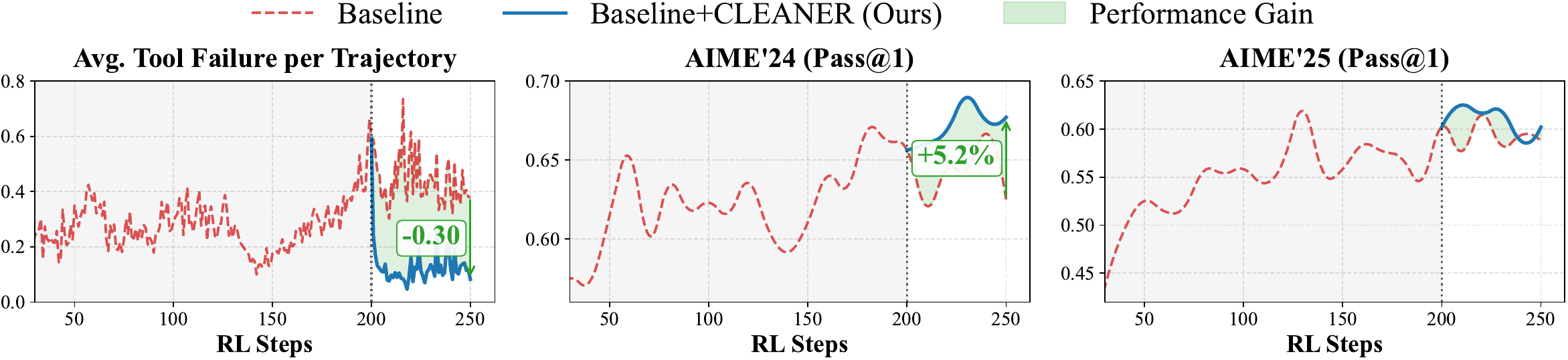}
    \caption{\textbf{Recovery from suboptimal policies.} Comparison of training metrics before and after introducing \method{} at step 200. The inclusion of \method{} effectively stabilizes the optimization process, leading to a marked improvement in final performance.}
    \label{fig:exp_repair}
\end{figure*}

\noindent \textbf{Internalization vs. Scaffolding.} To verify that \method{} effectively internalizes reasoning patterns, we evaluate the model with the SAAR mechanism deactivated. As detailed in Table~\ref{tab:exp_ablation_w_o_saar}, the model retains robust performance even in the absence of this ``scaffolding.'' Specifically, on AIME24, Pass@1 and Pass@16 decline by only 0.6\% and 1.3\%, respectively; on AIME25, Pass@1 decreases by 2.5\%, while Pass@16 exhibits a marginal gain of 0.2\%. Similarly, GPQA and LiveCodeBench show negligible performance degradation. This confirms that the policy has assimilated the error-avoidance logic into its intrinsic parameters, enabling efficient deployment without external dependencies.
Alternatively, SAAR can serve as a lightweight inference enhancement. It incurs a mere 8.8\% increase in average latency—significantly lower than computational-heavy test-time scaling methods like tree search~\citep{bi2024forest,yao2023tree}—while effectively mitigating in-context code errors and improving stability.

\noindent \textbf{Recovery from Suboptimal Policies.} To assess the restorative capability of our method, we performed a recovery experiment initializing from step 200 of the DAPO-baseline. At this time, the baseline exhibited significant instability, averaging 0.6 erroneous tool invocations per trajectory. As shown in Fig.~\ref{fig:exp_repair}, \method{} quickly stabilized training, effectively suppressing erroneous invocations and reducing the average number of tool calls per trajectory. Consequently, accuracy on AIME24 and AIME25 improved by 5.2\% and 1.0\%, respectively. However, we observe that this post-hoc recovery failed to reach parity with models trained with SAAR from scratch, underscoring the necessity of integrating the mechanism throughout the entire training lifecycle.

\section{Conclusion}
To address the inefficiency in agentic RL caused by execution noise and ambiguous credit assignment. We propose {\method{}}, which employs {Similarity-Aware Adaptive Rollback} to transform noisy exploration logs into clean, self-purified trajectories prior to optimization. This aligns training signals with correct behavior, enabling models to internalize robust tool usage without error interference. Empirical results on AIME24/25, GPQA, and LiveCodeBench show average accuracy gains of 6\%, 3\%, and 5\% over baselines. Crucially, \method{} matches the performance of SOTA methods while requiring only one-third of the training steps, highlighting trajectory purification as a scalable alternative for efficient agentic RL.

\bibliography{iclr2026_malgai}
\bibliographystyle{iclr2026_conference}
\newpage
\appendix
\onecolumn
\section{Detail Preliminaries}~\label{app:grpo}
To optimize the policy efficiently without the overhead of a value function critic, we operate within the \textbf{Group Relative Policy Optimization (GRPO) framework}~\cite{guo2025deepseekr1}. This paradigm estimates the baseline from group statistics rather than a separate neural network. Specifically, for each query $q$, a group of $G$ trajectories $\{\tau_i\}_{i=1}^G$ is sampled from the current policy $\pi_{\theta_{old}}$. The advantage $A_i$ for the $i$-th trajectory is derived by normalizing its reward against the group statistics:
\begin{equation}
A_i = \frac{R(\tau_i) - \mu_R}{\sigma_R + \delta},
\end{equation}
where $\mu_R$ and $\sigma_R$ denote the mean and standard deviation of the group rewards, respectively. Following this formulation, the policy is updated by maximizing the surrogate objective:
\begin{equation}
\begin{split}
\mathcal{J}(\theta) = \mathbb{E}_{\substack{q \sim P(Q) \\ \tau \sim \pi_{\theta_{old}}}} \bigg[ \frac{1}{G} \sum_{i=1}^G \min \Big( & \rho_i A_i, \\
& \text{clip}(\rho_i, 1-\epsilon, 1+\epsilon) A_i \Big) \bigg],
\end{split}
\end{equation}
where $\rho_i = \frac{\pi_\theta(\tau_i|q)}{\pi_{\theta_{old}}(\tau_i|q)}$ represents the importance sampling ratio, and $\epsilon$ is the clipping hyperparameter. This objective serves as the optimization backbone for our training process.

\section{Implementation Details}~\label{app:hyper}
\begin{table}[!h]
    \centering
    \begin{minipage}{0.58\linewidth}
        \centering
        \caption{Hyperparameters for Reinforcement Learning.}
        \label{tab:hyperparameters}
        \resizebox{\linewidth}{!}{ 
        \begin{tabular}{lc}
            \toprule
            \textbf{Hyperparameter} & \textbf{Value} \\
            \midrule
            Learning Rate & $2 \times 10^{-6}$ (4B) / $1 \times 10^{-6}$ (7B) \\
            Max Prompt Length & $2{,}560$ \\
            Max Response Length & $20{,}480$ (Avg. $\approx 7{,}000$) \\
            LR Warmup Steps & $20$ \\
            PPO Clip Ratio ($\epsilon^{-},\epsilon^+$) & $0.20,0.28$ \\
            \midrule
            Retry Limit $K$ & $3$ \\
            Similarity Threshold $\gamma$ & $0.5$ \\
            Reward Type & Outcome-only $\{-1, 1\}$ \\
            \bottomrule
        \end{tabular}
        }
    \end{minipage}
    \hfill 
    \begin{minipage}{0.34\linewidth}
        \centering
        \normalsize
        \caption{Sampling configurations for evaluation.}
        \label{tab:sampling}
        \resizebox{\linewidth}{!}{ 
        \begin{tabular}{lc} 
            \toprule
            \textbf{Hyperparameter} & \textbf{Value} \\
            \midrule
            Temperature & $1.0$ \\
            Top-$p$ & $0.6$ \\
            Top-$k$ & $-1$ \\
            \bottomrule
        \end{tabular}
        }
    \end{minipage}
\end{table}
\paragraph{Models and Training Datasets}
For the RL phase, we utilize the open-source dataset from \citep{yu2025demystifying} which comprises a diverse mixture of 17k samples from DAPO-Math~\citep{yu2025dapo}, 4,902 math and 3,586 code samples from Skywork-or1~\citep{he2025skywork}, and 3k science problems from MegaScience~\citep{fan2025megascience}.
\paragraph{Training Configurations}
Table~\ref{tab:hyperparameters} summarizes the hyperparameters for our reinforcement learning stage. We employ different learning rates for models of varying scales: $2 \times 10^{-6}$ for the 4B model and $1 \times 10^{-6}$ for the 7B model. Although the maximum context window is set to $20{,}480$ tokens to accommodate long-horizon trajectories, the empirical average sequence length across our training data remains approximately $7{,}000$ tokens, ensuring computational efficiency without sacrificing context.

\paragraph{Ablation on Key Parameters}
The retry limit $K$ and similarity threshold $\gamma$ are critical for the \method{} framework. Through empirical validation, we found that $K=3$ offers an optimal balance between recovery rate and computational cost; values below $3$ lead to a noticeable drop in the recovery of successful trajectories, while values above $3$ yield diminishing returns. Regarding the similarity threshold $\gamma$, our method exhibits strong robustness across a range of values. However, we recommend a relatively high threshold (between $0.5$ and $1.0$) to ensure high-fidelity trajectory refinement, with $0.5$ being our default setting.
\paragraph{Hardware and Compute Costs}
All experiments were conducted on a single node equipped with 4$\times$NVIDIA H100 or 4$\times$NVIDIA H200 GPUs. For the Qwen3-4B model, the full training cycle takes approximately 4 days. Interestingly, training the Qwen2.5-7B model requires only 2 days. This is primarily due to our data filter process, where we filter out both overly simplistic and excessively difficult samples to focus the training on more informative trajectories.

\paragraph{Evaluation Sampling Parameters}
Table~\ref{tab:sampling} lists the specific decoding configurations used during the evaluation phase to ensure consistent performance comparison.

\section{Unsuccessful Attempts: Leveraging Negative Samples via SAAR}

During the development of the \textbf{\method{}} framework, we explored an alternative strategy to further enhance model performance by utilizing the erroneous actions identified by the \textbf{SAAR} mechanism as negative samples. Despite investigating multiple configurations, these attempts did not yield the expected improvements. We share these findings here to provide insights for future research in agentic RL.

\paragraph{Motivation} 
The \textbf{SAAR} mechanism primarily functions by overwriting failed tool invocations with correct actions, ensuring that the agent is exposed to high-quality, "purified" trajectories during training. We hypothesized that the original erroneous actions, which are discarded in the standard \method{} pipeline, could serve as valuable negative signals. By explicitly learning what constitutes an incorrect behavior through contrastive signals, the model might further refine its coding and tool-use capabilities.

\paragraph{Implementation} 
For each successful trajectory recovery via \textbf{SAAR}, we paired the original failed tool call with the corrected rollout to generate online positive-negative pairs. We then applied an \textbf{online-DPO} (Direct Preference Optimization) objective to these pairs. To isolate the impact of the tool call itself, we masked all tokens except for the tool invocation segment, penalizing the failed attempt while rewarding the corrected one. During optimization, the DPO gradient was integrated with the GRPO signal at each step to perform a unified policy update.

\paragraph{Results and Analysis} 
While this approach slightly improved the model's ability to "self-repair" specific code snippets, it failed to improve the overall success rate on complex reasoning tasks and even triggered training collapse in the later stages. Our analysis suggests several reasons for this failure:

\begin{enumerate}[9]
    \item[\ding{182}] \textbf{Reasoning vs. Syntactic Failure:} Tool invocation failures in advanced agents are frequently rooted in faulty reasoning rather than mere syntax errors. As training progresses, the \textbf{SAAR} mechanism shifts from addressing minor typos toward "deep" logical repairs. Penalizing only the tool call segment without addressing the preceding CoT creates a disconnect between the agent's internal logic and its external actions. This imbalance may discourage the generation of complex code instead of fostering better reasoning.
    \item[\ding{183}] \textbf{Correlation between Reasoning and Code Proficiency:} Code-use proficiency is intrinsically linked to the agent's overall reasoning capacity. We observed that as the model's reasoning improves, it naturally adopts more sophisticated code structures. Relying on simple token-level masking to reward/penalize specific segments can be counterproductive. Furthermore, recent studies (e.g., \textbf{GSPO}~\citep{zheng2025groupspo}) indicate that assigning disproportionate weights to specific tokens within a single trajectory can adversely affect policy optimization stability.
\end{enumerate}
Ultimately, effectively utilizing negative samples and determining their net value in agentic RL remains an open question for future study.
\section{Related Work}

\paragraph{Static and Supervised Tool-Integrated Reasoning.}
Tool-integrated reasoning (TIR) empowers LLMs to offload precise computations to external environments~\citep{parisi2022talm,schick2023toolformer, wang2024executable}. Foundational paradigms like ReAct~\citep{yao2022react} and Program of Thoughts~\citep{chen2022program} established the viability of interleaving reasoning with execution. Scaling these capabilities, recent Supervised Fine-Tuning (SFT) approaches~\citep{gou2023tora,qin2023toolllm,schick2023toolformer} and unified executable frameworks like CodeAct~\citep{wang2024executable} have achieved remarkable performance by mimicking expert trajectories. 
However, relying solely on behavioral cloning limits models to successful demonstrations. Consequently, these agents often mimic surface-level patterns without grasping the underlying causality, leaving them ill-equipped to handle the inherent noise of real-world tool interactions.~\citep{wang2024trove, kumar2024training}.

\paragraph{Agentic RL.}
To bridge the gap left by supervised methods, Agentic RL treats tool invocation (e.g., executable code~\citep{wang2024executable}) as an explicit action space, optimizing adaptive strategies via outcome-driven rewards~\citep{shridhar2020alfworld, mialon2024gaia}. This paradigm enables agents to move beyond simple imitation toward discovering flexible solutions in open-ended tasks~\citep{tan2024true, bai2024digirl, wang2024distrl}. Recent advancements have systematically scaled these capabilities to autonomous search and query refinement~\citep{jin2025search, song2025r1, sun2025zerosearch}, long-horizon research tasks~\citep{li2025webthinker, li2025websailor}, and complex multi-tool coordination~\citep{singh2025agentic, dong2025toolstar, qian2025toolrl, wang2025otc, wang2025ragen}. These developments are further supported by studies on scaling laws~\citep{li2025torl} and the strategic logic of tool invocation~\citep{feng2025retool}. 
Crucially, recent research has begun to prioritize fundamental training stability and exploration efficiency. While \textit{Demystifying RL}~\citep{yu2025demystifying} investigates fundamental training recipes and \textit{rStar2-Agent}~\citep{shang2025rstar2} mitigates execution noise via trajectory filtering, ARPO~\citep{dong2025arpo} and AEPO~\citep{dong2025aepo} specifically focus on enhancing exploration. They introduce entropy-regulated mechanisms to dynamically modulate rollouts, leveraging model uncertainty to improve performance in multi-turn interactions. Despite these improvements, existing methods still struggle to effectively decouple high-quality signals from the pervasive noise inherent in complex tool-use trajectories, which often leads to sub-optimal policy updates. To address this, our \method{} framework introduces a robust mechanism to refine training data and stabilize the learning process.

\end{document}